\documentclass[sigconf, screen]{acmart}
\setcopyright{none}
\renewcommand\footnotetextcopyrightpermission[1]{}
\AtBeginDocument{%
  }

\setcopyright{acmlicensed}
\copyrightyear{2026}
\acmYear{2026}
\acmDOI{XXXXXXX.XXXXXXX}
\AtBeginDocument{%
  \fancyhead[LE]{Preprint}%
  \fancyhead[RO]{Preprint}%
}
\acmISBN{978-1-4503-XXXX-X/2018/06}

\usepackage{enumitem}
\usepackage{multirow}  
\usepackage{booktabs}  
\usepackage{graphicx}  
\usepackage{wrapfig}
\usepackage{xcolor}
\usepackage{tabularx}
\newcolumntype{L}{>{\raggedright\arraybackslash}X}

\begin{document}

\title{Foundation Models Meet Agriculture: Challenges Beyond Pretraining}

\author{Vishal Nedungadi}
\orcid{0009-0008-3054-7468}
\correspondingauthor
\affiliation{%
  \institution{Artificial Intelligence and Plant Science Group}
  \institution{Wageningen University and Research}
  \city{Wageningen}
  \country{Netherlands}
}
\email{vishal.nedungadi@wur.nl}

\author{Xingguo Xiong}
\affiliation{%
  \institution{Institute of Digital Agriculture}
  \institution{Zhejiang Academy of Agricultural Sciences}
  \city{Hangzhou}
  \country{China}
}
\email{xiong\_xg@zju.edu.cn}

\author{Marc Ru\ss{}wurm}
\affiliation{%
 \institution{Machine Learning in Earth Observation Lab}
 \institution{University of Bonn}
 \city{Bonn}
 \country{Germany}}
\email{marc.russwurm@uni-bonn.de}

\author{Ioannis N. Athanasiadis}
\orcid{0000-0003-2764-0078}
\affiliation{%
  \institution{Artificial Intelligence and Plant Science Group}
  \institution{Wageningen University and Research}
  \city{Wageningen}
  \country{Netherlands}}
\email{ioannis.athanasiadis@wur.nl}

\renewcommand{\shortauthors}{Nedungadi et al.}

\begin{abstract}
Global food security and sustainable climate action increasingly rely on robust, scalable agricultural monitoring. Earth observation foundation models have emerged as powerful, label-efficient tools across general remote sensing domains, yet early attempts to deploy them for agricultural applications have yielded surprisingly poor results. We hypothesize that this performance gap stems from the extreme heterogeneity of agricultural landscapes and the inherent inability of current earth observation foundation models to adapt to task-specific nuances. In this work, we systematically evaluate two critical bottlenecks hindering the deployment of foundation models in agricultural tasks, benchmarking two earth observation foundation models, a foundation model designed for tabular data, and conventional supervised baselines across seven real-world agricultural datasets spanning yield prediction, phenology estimation, and crop classification. First, we identify a pretraining-deployment modality gap: agricultural downstream tasks frequently require diverse, non-imagery data modalities that earth observation foundation models are architecturally unequipped to ingest, while a foundation model built for tabular data handles this heterogeneity more naturally. Second, we formalize the agricultural task space across five structural axes to demonstrate why current models fail to generalize reliably, resulting in highly unstable model rankings across evaluation settings. By characterizing these structural and modal gaps, our insights highlight the friction between general-purpose architectures and specialized agricultural downstream data, providing a strategic roadmap for developing the next generation of domain-aware foundation models.
\end{abstract}



\begin{teaserfigure}
 \centering
 \includegraphics[width=0.9\linewidth]{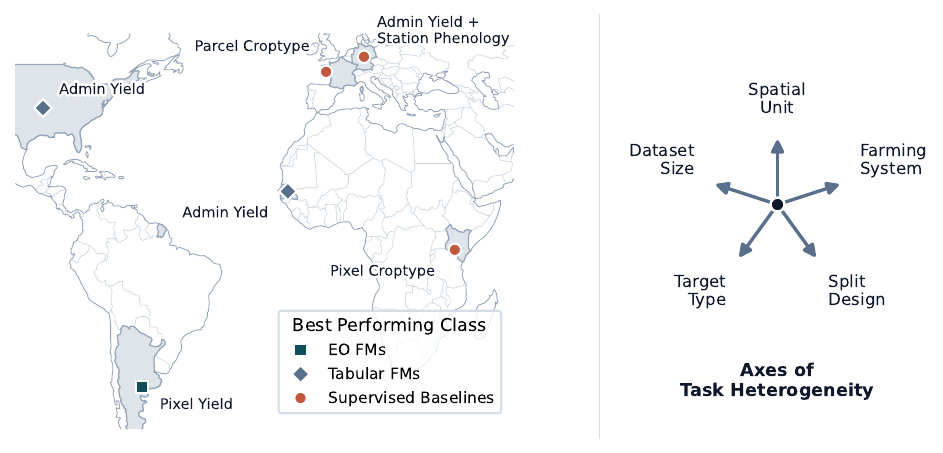}
  \caption{The map (left) demonstrates the geographic diversity of our evaluation and highlights the winning model family for each region. The absence of a universal winner is driven by the structural complexity of the tasks, which vary drastically across five key dimensions of shifts (right).}
  \label{fig:teaser}
\end{teaserfigure}


\maketitle

\section{Introduction}

Agricultural monitoring has evolved from simple economic utility into a critical instrument for global food security. As climate change intensifies extreme weather events, the historical baselines that once made crop production predictable are rapidly destabilizing \cite{doi:10.1126/science.1204531, IntergovernmentalPanelonClimateChangeIPCC2023}. Managing this risk depends on monitoring systems that are scalable, timely, and precise across highly diverse geographies, and that together answer three interdependent questions central to food security: \emph{where} crops are grown (crop type mapping), \emph{when} they develop (phenology estimation), and \emph{how much} they will produce (yield estimation). Although these tasks draw on the same underlying biophysical signals (vegetation and soil state, and meteorological forcing), building supervised models for them at a global scale remains fundamentally bottlenecked by labels; dense, high-quality ground truth is expensive and logistically difficult to acquire across the full range of climates, farming systems, and topographies where such systems must operate \cite{Ma2024,Mohammadi2024,Wang2023}.

Foundation models (FMs) have been proposed to relieve exactly this bottleneck, amongst others. In the broader remote sensing domain, Earth Observation (EO) FMs learn generalized spatial and temporal representations by self-supervised pretraining on massive archives of unlabeled multi-modal satellite imagery \cite{tseng2024lightweightpretrainedtransformersremote, tseng2025galileolearningglobal, nedungadi2024mmearthexploringmultimodalpretext, brown2025alphaearthfoundationsembeddingfield, feng2026tesseratemporalembeddingssurface}. Because agricultural tasks share many of these same satellite modalities, the pretrained priors are expected to transfer with only minimal labeled data \cite{jakubik2023foundationmodelsgeneralistgeospatial, jakubik2026terramindlargescalegenerativemultimodality, Qin2025,Xu2024}. Throughout this work we are concerned specifically with this transfer setting, in which \emph{pretrained} FMs are adapted to downstream tasks \cite{Athanasiadis2026}.

In practice, however, this promise has proven fragile. General EO benchmarks report that no single pretrained backbone dominates and that model rankings shift with sensor, resolution, and task \cite{marsocci2025pangaeaglobalinclusivebenchmark, simumba2026geobench2performancecapabilityrethinking}, and within agriculture the same backbones transfer unevenly even across regions of a single task \cite{shang2026benchmarkinggeospatialfoundationmodels}. We argue that this instability is not merely a symptom of geographic distribution shift, but a deeper mismatch between what a general-purpose EO FM assumes and what agricultural monitoring actually requires. Even as EO FMs have grown more flexible, ingesting multiple modalities and operating across spatial scales \cite{tseng2024lightweightpretrainedtransformersremote, tseng2025galileolearningglobal, xiong2025neuralplasticityinspiredmultimodalfoundation}, each model still fixes, at pretraining time, both the set of modalities it can represent and the representational structure it imposes on its inputs, while agriculture routinely steps outside both. 
On the input side, agricultural tasks hinge on modalities that are not imagery at all, such as deep soil properties, agro-meteorological indices, and daily weather \cite{kamangir2025californiacropyieldbenchmark,essd-18-3997-2026}. Furthermore, each task can be realized through many possible datasets that often considers a different subset of these modalities. Therefore no fixed pretraining modality set covers every task a deployed FM will face; we refer to this as \emph{modality mismatch}. 
On the task side, agricultural monitoring is not a single problem but many, and even datasets within the same task can differ in spatial unit, temporal framing, label density, and farming system, such that no single pretraining objective can be well-aligned to all of them at once; we refer to this as \emph{task heterogeneity}. These two challenges organize our analysis.

To make these two challenges measurable rather than anecdotal, we compare three model families that differ along two axes: whether a model carries pretrained priors, and whether it can natively ingest arbitrary modalities. \textit{EO FMs} carry priors but are modality-constrained; \textit{supervised baselines} trained from scratch carry none but ingest whatever is available; \textit{tabular FMs} have both, though their priors can come from massive synthetic tabular data rather than satellite imagery \cite{Hollmann2025}. This last combination is pivotal, as it lets us separate the cost of the modality mismatch from the value of pretraining itself. We evaluate these families across seven representative, real-world datasets spanning three task and a wide range of geographies and farming systems, from industrial agriculture in the US and Europe to smallholder systems in Senegal and Kenya.

Across this suite, both challenges surface clearly: EO FMs incur a substantial penalty whenever a task depends on modalities outside their pretraining set, and even when no mismatch exists, no single architecture ranks consistently. Rather than averaging this instability away, we diagnose why current FMs underperform in agriculture and chart what domain-aware models must address. Concretely, our contributions are as follows:

\begin{itemize}[leftmargin=0.35cm, labelsep=0.15cm]
    \item \textbf{Characterizing the modality mismatch.} We quantify the modality mismatch as the fraction of each task's available modalities an EO FM can natively ingest, and test whether pretraining compensates for what is missing.
    \item \textbf{Formalizing task heterogeneity.} We characterize the agricultural task space along five structural axes (spatial unit, farming system, split design, task type, and dataset size) and reflect on if this heterogeneity results in unstable ranking of models.
    \item \textbf{Implications for agricultural FM design.} We show that a domain-agnostic tabular FM, despite carrying no geospatial or agricultural priors, is competitive with or superior to EO FMs in several settings. We distill our results into open challenges for the community.
\end{itemize}

\section{Experimental Setup}
\subsection{The Task Suite}
We identify Crop Type Mapping, Crop Phenology Estimation, and Crop Yield Estimation as three representative \emph{tasks} for addressing global food security challenges \cite{nedungadi2025generalspecializedneedfoundational}. Among these, we curate seven highly diverse \emph{datasets} spanning five structural axes (top panel in Figure \ref{fig:performance_heatmap}), including spatial unit, farming system, split design, target type and dataset size. A description of each dataset is provided below. The specific modalities curated for each task are detailed in the top partition of Table \ref{tab:modality_coverage}. We note that several of the underlying benchmarks and tasks cover many more regions than we evaluate here, and our chosen regions are not necessarily representative of the full diversity within each broader benchmark. Region selection was instead driven by the goal of axis coverage together with data quality and availability. Nevertheless, we believe the resulting suite is representative enough of the broader structural variation at stake and to support the conclusions about task heterogeneity we discuss in Section~\ref{sec:task_heterogeneity}.

\begin{itemize}[leftmargin=0.35cm, labelsep=0.15cm]
    \item CropHarvest \cite{tseng2021cropharvest}: A pixel-level crop type mapping dataset with monthly temporal steps. While the original dataset is global, we explicitly isolate the \textit{Kenya} subset to stress-test models on a fragmented, smallholder farming system. Similar to \cite{tseng2025galileolearningglobal}, we formulate this as a Maize vs. Rest binary classification problem, noting that the task is highly class-imbalanced.
    
    \item Breizhcrops \cite{breizhcrops2020}: A parcel-level 9-class crop classification dataset situated in \textit{Brittany, France}. It represents a large-scale farming system monitored with dense, high-frequency optical time series (45 temporal steps).
    
    \item BloomBench \cite{bree2026bloombench}: A phenology benchmark on multi species fruit trees. We adapt the original benchmarks evaluation protocols, data splits, and preprocessing code to model maize in \textit{Germany}. The problem is formulated as a time-to-event regression task to estimate the date of tassel emergence. Ground truth phenological labels are sourced from the PEP725 database, while the predictive modalities consist exclusively of daily meteorological forcing (365 time steps) derived from ERA5 reanalysis data \cite{Hersbach2020}. 
    
    \item CY-Bench \cite{essd-18-3997-2026}: A sub-national level crop yield estimation benchmark covering maize and wheat in 30 countries. We choose \textit{United States, Germany and Senegal}, representing both industrial and smallholder farming systems. Unlike pixel-level tasks, this dataset predicts heavily aggregated administrative yields (e.g., counties or regions, depending on the country). 
    
    \item YieldSAT \cite{miranda_2026_yieldsat}: A massive-scale, pixel-level corn yield estimation benchmark covering 4 countries, where ground-truth labels are derived directly from combine harvester yield monitors. We select \textit{Argentina}, and use the original datasets input modalities that are processed into a fixed 24 timesteps based on the sentinel-2 coverage.
\end{itemize}

\subsection{Models}
We evaluate models across three families: \textit{Supervised Baselines}, \textit{EO FMs} and \textit{Tabular FMs}.

\paragraph{\textbf{Supervised Baselines}} We include supervised baselines because they represent the class of models with no architectural constraint on input modalities and can be freely designed around each task's available features. We consider a \textit{Random Forest}, a strong, well-established baseline for tabular and concatenated feature sets; an \textit{LSTM}, which explicitly models temporal dependencies such as phenological and seasonal weather trends across the time-series inputs; and a \textit{Transformer}, trained from scratch, which isolates the raw benefit of the attention mechanism on each task's temporal dynamics independent of any large-scale pretraining. We intentionally select these over highly engineered architectures to minimize architectural inductive biases, allowing us to focus purely on the complexity of the task \cite{grinsztajn2022treebasedmodelsoutperformdeep, Paudel2021, Jeong2016}. 

\paragraph{\textbf{EO FMs}} For EO FMs, several pretrained backbones exist \cite{tseng2024lightweightpretrainedtransformersremote, tseng2025galileolearningglobal, jakubik2023foundationmodelsgeneralistgeospatial, jakubik2026terramindlargescalegenerativemultimodality}; we focus on models that operate over pixel time series, since this representation aligns naturally with the structure of our datasets, and select \textit{Galileo} \cite{tseng2025galileolearningglobal} as a strong, actively maintained representative of this class. To probe whether a more accommodating prior could close this gap, we additionally instantiate \emph{CropFM}, an idealized, purpose-built EO FM. We build CropFM as an extension of the lightweight Presto architecture \cite{tseng2024lightweightpretrainedtransformersremote} while making it natively ingest more modalities and finer (weekly) timesteps. We deliberately exclude precomputed global embedding products such as AlphaEarth \cite{brown2025alphaearthfoundationsembeddingfield} and Tessera \cite{feng2026tesseratemporalembeddingssurface} as their embeddings are precomputed at a temporal granularity misaligned with our datasets; we highlight that there needs to be a dedicated study on using these embeddings for such tasks. 

\paragraph{\textbf{Tabular FMs}} Finally, distinct from both of these families, tabular FMs carry generic priors unrelated to EO or agriculture, while imposing no modality constraints whatsoever; we include \textit{TabPFN} \cite{Hollmann2025}, pretrained on large-scale synthetic data via in-context learning, to represent this family and test whether such domain-agnostic priors transfer to agricultural monitoring.

\begin{table*}[t!]
\centering
\caption{Modality Coverage Matrix. The top partition shows the input modalities curated for each dataset. The bottom partition shows the physical modalities supported natively by each model architecture. Acronyms: S1/S2 (Sentinel-1/2), NDVI (Normalized Difference Vegetation Index), FAPAR (Fraction of Absorbed Photosynthetically Active Radiation), DEM (Digital Elevation Model), Basic Meteo (standard weather variables like temperature and precipitation), Ag-Meteo (agro-meteorological variables such as radiation and climatic water balance), and SSM (Surface Soil Moisture).}
\label{tab:modality_coverage}
\begin{tabular}{l cc cc c cc cc}
\toprule
 & \multicolumn{2}{c}{\textbf{Primary EO}} & \multicolumn{2}{c}{\textbf{Derived}} & \textbf{Topo.} & \multicolumn{2}{c}{\textbf{Meteo.}} & \multicolumn{2}{c}{\textbf{Subsurface}} \\
\cmidrule(lr){2-3} \cmidrule(lr){4-5} \cmidrule(lr){6-6} \cmidrule(lr){7-8} \cmidrule(lr){9-10}
\textbf{Task / Model} & \textit{S2} & \textit{S1} & \textit{NDVI} & \textit{FAPAR} & \textit{DEM} & \textit{Basic} & \textit{Ag-Meteo} & \textit{SSM} & \textit{Deep Soil} \\
\midrule
\multicolumn{10}{l}{\textbf{\textit{Downstream Datasets (Available Modalities)}}} \\
\addlinespace
\multicolumn{10}{l}{\textbf{Yield Estimation}} \\
\quad CYBench (Germany)     & $\bullet$ & $\bullet$ & $\bullet$   & $\bullet$   & $\circ$   & $\bullet$ & $\bullet$ & $\bullet$ & $\circ$   \\
\quad CYBench (US)          & $\bullet$ & $\bullet$ & $\bullet$   & $\bullet$   & $\circ$   & $\bullet$ & $\bullet$ & $\bullet$ & $\circ$   \\
\quad CYBench (Senegal)     & $\bullet$ & $\bullet$ & $\bullet$   & $\bullet$   & $\circ$   & $\bullet$ & $\bullet$ & $\bullet$ & $\circ$   \\
\quad YieldSat (Argentina)  & $\bullet$ & $\circ$ & $\circ$   & $\circ$   & $\bullet$   & $\bullet$ & $\circ$ & $\circ$   & $\bullet$ \\
\addlinespace
\multicolumn{10}{l}{\textbf{Phenology Estimation}} \\
\quad BloomBench (Germany)   & $\circ$ & $\circ$   & $\circ$   & $\circ$   & $\circ$   & $\bullet$ & $\circ$   & $\circ$   & $\circ$   \\
\addlinespace
\multicolumn{10}{l}{\textbf{Crop Type Mapping}} \\
\quad CropHarvest (Kenya)   & $\bullet$ & $\circ$   & $\bullet$   & $\circ$   & $\bullet$ & $\bullet$   & $\circ$   & $\circ$   & $\circ$   \\
\quad BreizhCrop (France)   & $\bullet$ & $\circ$   & $\circ$ & $\circ$   & $\circ$   & $\circ$   & $\circ$   & $\circ$   & $\circ$   \\
\midrule
\multicolumn{10}{l}{\textbf{\textit{Model Architectures (Natively Supported)}}} \\
\addlinespace
CropFM                & $\bullet$ & $\bullet$ & $\circ$   & $\bullet$   & $\bullet$   & $\bullet$   & $\bullet$   & $\circ$   & $\bullet$   \\
Galileo               & $\bullet$ & $\bullet$ & $\bullet$   & $\circ$   & $\bullet$ & $\bullet$ & $\circ$   & $\circ$   & $\circ$   \\
Supervised / TabPFN        & $\bullet$ & $\bullet$ & $\bullet$ & $\bullet$ & $\bullet$ & $\bullet$ & $\bullet$ & $\bullet$ & $\bullet$ \\
\bottomrule
\end{tabular}
\end{table*}

\begin{figure*}
    \centering
    \includegraphics[width=\linewidth]{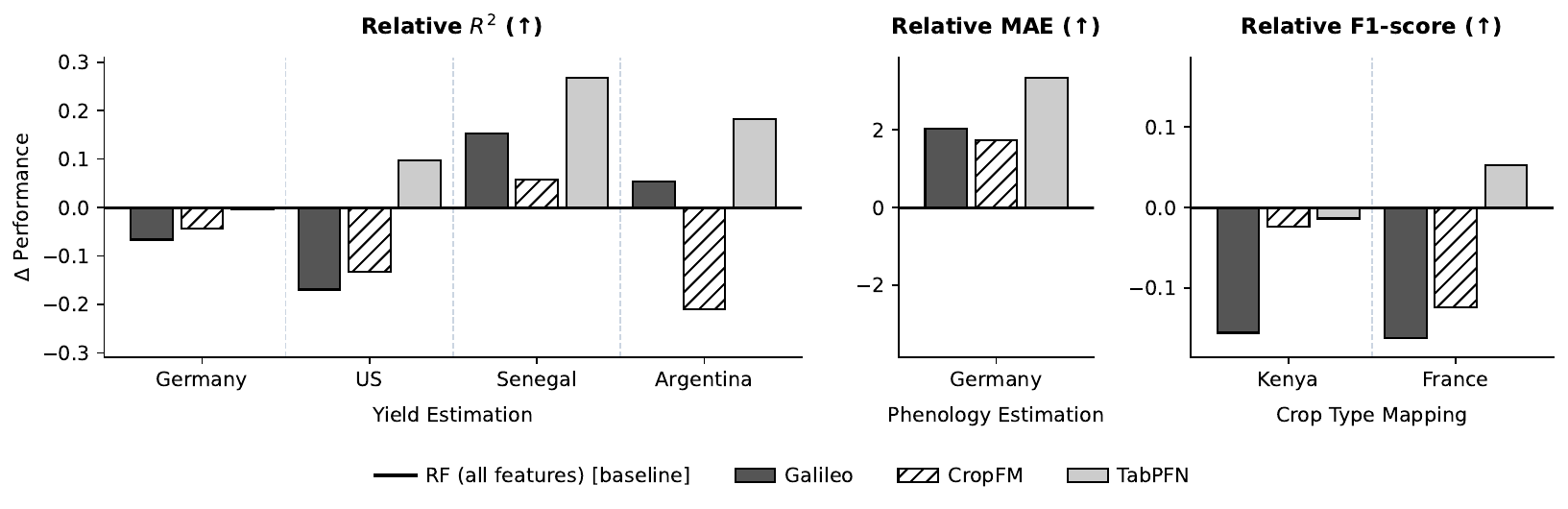}
    \caption{\textbf{Relative Performance Gap.} The zero-line represents the unconstrained Random Forest baseline utilizing all available dataset specific modalities. Bars show the relative performance ($\Delta$) of models restricted to their natively supported modalities and evaluated using frozen features and a Random Forest. To ensure positive values always indicate improvement, $\Delta$ is computed as $\text{Model} - \text{Baseline}$ for $R^2$ and F1-score, and $\text{Baseline} - \text{Model}$ for MAE. Negative values explicitly highlight the performance gap incurred from missing dataset-critical modalities. }
    \label{fig:relative_modality_gap}
\end{figure*}

\subsection{Evaluation}

\paragraph{\textbf{Task Metrics}}

Since our evaluation spans fundamentally different tasks (and datasets), we evaluate performance using dataset-specific metrics. For classification datasets (BreizhCrop and CropHarvest), we report the Macro F1-score. For regression datasets (Phenology, YieldSAT, and CY-Bench), we report the Mean Absolute Error (MAE) for phenology, which quantifies the average prediction error in days, and the coefficient of determination ($R^2$) for yield estimation, measuring the proportion of variance explained by the model.

\paragraph{\textbf{Evaluation Splits}}
To evaluate model generalization across different dimensions, we apply tailored data splitting strategies suited to each dataset. For CropHarvest, we employ a standard 70/15/15 randomized pixel split, whereas Breizhcrops follows a strict spatial holdout based on its original evaluation protocol to test geographic generalization. To assess temporal extrapolation, both Phenology estimation and YieldSAT utilize chronological holdouts; Phenology estimation relies on a 70/15/15 chronological split to evaluate on unseen future weather patterns, while YieldSAT reserves the final available year as a strict temporal test set. Finally, to simulate operational forecasting for CY-Bench, we employ a chronological walk-forward validation strategy on the final 5 years. Instead of a single static test set, the model iteratively trains on all available historical data up to year $t$ to predict yields for the unseen year $t+1$. In the subsequent iteration, year $t+1$ is incorporated into the training set to predict year $t+2$. 

\paragraph{\textbf{Evaluation Protocol}}

All experiments are deterministic runs repeated with three random seeds, and we report the mean performance across runs. Standard deviations for all results are provided in Appendix \ref{sec:app_full_results}.

For the EO FMs (CropFM and Galileo),we evaluate two adaptation strategies representing  two dominant paradigms for reusing pretrained representations: (i) full fine-tuning, which adapts the underlying representation itself, and (ii) frozen feature extraction, in which a Random Forest is trained on the frozen FM embeddings, which instead probes the representation as-is. For full fine-tuning, we adopt a two-stage warm-up schedule to mitigate the risk of pretrained features being distorted by an initially unadapted head \cite{kumar2022finetuningdistortpretrainedfeatures}: we first optimize only the task-specific prediction head for five epochs using a higher learning rate ($1e-4$), after which all model parameters are jointly optimized for an additional 25 epochs using a lower learning rate ($1e-3)$. The LSTM and Transformer baselines are also trained for 30 epochs. Finally, we emphasize that all models only accept the inputs that they are capable of accepting. 

\paragraph{\textbf{Temporal Aggregation}}

For each task, we define a standard temporal resolution used consistently across Random Forest, LSTM, Transformer, TabPFN, and CropFM. Where a task's raw data is available at high or irregular frequency, we aggregate it to a more tractable resolution: BreizhCrop's individual Sentinel-2 acquisitions are aggregated to weekly composites, CY-Bench's high-frequency meteorological and vegetation index observations are similarly aggregated to weekly, and Phenology's daily meteorological observations are also aggregated to weekly. CropHarvest and YieldSAT require no further aggregation.

Galileo imposes an additional constraint beyond this standard resolution: its fixed maximum sequence length is exceeded by the weekly resolution used for BreizhCrop, CY-Bench, and Phenology. For these three tasks only, we aggregate further to monthly composites specifically for Galileo, while CropHarvest and YieldSAT already fall within its sequence budget at the standard resolution and are left unchanged.

\paragraph{\textbf{Hyperparameter Policy}}

We adopt a standardized set of hyperparameters across datasets. This design reflects two considerations. First, several datasets contain only a limited number of labeled samples, making extensive hyperparameter optimization unreliable and prone to overfitting. Second, our objective is to evaluate model robustness under a realistic benchmarking protocol instead of measuring the best achievable performance on each individual dataset. A fixed evaluation protocol also enables a fairer comparison across tasks with widely varying label availability and distribution shifts \cite{Franceschi_2025}.

\paragraph{\textbf{Model Configurations}}

The LSTM consists of two layers with a hidden dimension of 128 and a dropout rate of 0.2. The Transformer uses an embedding dimension of 128, two encoder layers, eight attention heads, a feed-forward dimension of 128, and a dropout rate of 0.1. The Random Forest classifier/regressor is trained with 500 trees using the default settings for all remaining hyperparameters.

\section{Results and Empirical Analysis}
We structure our empirical evaluation around two primary structural challenges that govern model performance across the task suite: \textbf{Modality Mismatch} and \textbf{Task Heterogeneity}.

\subsection{Challenge 1: Modality Mismatch}
A fundamental bottleneck in adapting EO FMs to diverse agricultural tasks is the rigid architectural assumption regarding the input. Unlike supervised machine learning baselines that can dynamically ingest an arbitrary number of curated, heterogeneous modalities, pretrained architectures are strictly bounded by what is present in their pretraining distributions. This section breaks down the scale of this discrepancy and evaluates whether learned representations can overcome missing task-critical modalities.

\subsubsection{\textbf{The Pretraining-Downstream Modality Gap}}
\begin{figure}
    \centering
    \includegraphics[width=1.0\linewidth]{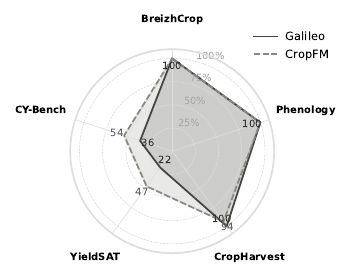}
    \caption{Modality Coverage Percentage. The radial chart illustrates the percentage of dataset-specific modalities natively supported by Galileo and CropFM.}
    \label{fig:radar_coverage}  
\end{figure}

We formally define the modality gap as the discrepancy between the rich modalities curated for specific datasets and the ingestion capabilities of a given FM. We quantify this gap using \textit{modality coverage}: the percentage of a dataset's curated modality set that a given model architecture can natively ingest, computed per dataset-model pair from the modality presence indicators in the Modality Coverage Matrix (Table \ref{tab:modality_coverage}; full computation details in Appendix Table~\ref{tab:granular_modalities}) and visualized in Figure \ref{fig:radar_coverage}). This gap is highly variable and dataset-dependent. Note that even when two datasets share a modality, the specific variables it comprises can differ, for example a 12-band Sentinel-2 acquisition versus a 10-band one, or a meteorological modality consisting of temperature alone versus temperature and precipitation together.

For crop classification datasets such as BreizhCrop, the required modalities (Sentinel-2 optical bands) align closely with the architectural expectations of CropFM and Galileo, resulting in near-complete modality coverage. Coverage narrows substantially, however, for datasets built around yield estimation, which depend on the deep soil and agro-meteorological signals we identified above as characteristic of the modality mismatch. Consequently, for datasets like CY-Bench and YieldSAT, true modality coverage drops significantly, challenging FMs' capacity to reconstruct meaningful embeddings when the modality mismatch is large.

\subsubsection{\textbf{Does Pretraining Compensate for Missing Modalities?}}

To determine if the rich latent representations learned during large-scale pretraining can compensate for the lack of explicit physical inputs, we analyze the relative performance ($\Delta$) between the constrained FM features (using a frozen feature extractor with a Random Forest head) and the unconstrained Random Forest baseline utilizing all available modalities (Figure \ref{fig:relative_modality_gap}), where a negative $\Delta$ indicates a performance penalty incurred during modality mismatch.

The results reveal that incomplete modality coverage imposes a highly context-dependent penalty, not a uniform one:
\begin{itemize}[nosep,leftmargin=0.35cm, labelsep=0.15cm]
    \item \textbf{Low Coverage Dominates Performance (Germany \& US):} For yield estimation in Germany and the US, both Galileo and CropFM underperform the baseline. Despite the depth of their pretraining, the models cannot compensate for missing critical physical modalities that the baseline utilizes natively.
    \item \textbf{The Volatility of Low Coverage (Argentina):} In the Argentina yield dataset, where native modality coverage is at its lowest (Table \ref{tab:modality_coverage}), we observe a stark architectural split. Galileo manages to yield a slightly positive $\Delta R^2 \approx 0.05$, whereas CropFM collapses to $\Delta R^2 \approx -0.21$. This divergence highlights that when critical physical modalities are missing, the reliability of the frozen representations becomes highly volatile and entirely dependent on how a specific model's pretraining objective happens to align with the target region.
    \item \textbf{High Coverage is No Guarantee (Kenya \& France):} Most surprisingly, in the crop type mapping tasks where modality coverage is near 100\% (Figure \ref{fig:radar_coverage}), the FMs still struggle. Galileo suffers performance penalties in both Kenya ($\Delta\text{F1} \approx -0.16$) and France ($\Delta\text{F1} \approx -0.16$). This indicates that while resolving modality mismatch is a necessary condition for robust agricultural modeling, it is not a sufficient one; even with perfect modality alignment, other localized factors disrupt frozen feature transfer (Section 4).
    \item \textbf{Benefit of Pretraining under Full Coverage (Germany Phenology):} In the German phenology task, where there is no modality mismatch (100\% coverage), the frozen FM features outperform the Random Forest baseline ($\Delta\text{MAE} > 0$).
\end{itemize}

\paragraph{\textbf{The Senegal Anomaly}} We note an exception in the Senegal yield estimation task, where both Galileo ($\Delta R^2 \approx +0.15$) and CropFM ($\Delta R^2 \approx +0.06$) produce positive deltas relative to the unconstrained RF baseline, despite operating under the same modality constraints as in Germany and the US. This result inverts the pattern observed before: rather than incomplete coverage dominating performance, the frozen FM representations prevent the downstream head from collapsing under extreme data scarcity. In this regime, we hypothesize that the unconstrained RF baseline lacks sufficient signal since Senegal represents smallholder farming systems.

\paragraph{\textbf{TabPFN and the Value of Unconstrained Feature Access}} A consistent finding across both challenges is the strong performance of TabPFN, a prior-data fitted network with no geospatial or agricultural pretraining. We find that it matches or outperforms FMs across several datasets. On Senegal specifically, TabPFN achieves a high positive delta ($\Delta R^2 \approx +0.22$). Unlike the other FMs, TabPFN operates without modality constraints: it natively ingests the full suite of available task features regardless of pretraining coverage, and applies a highly expressive prior over tabular distributions learned from synthetic data.

\begin{figure*}
    \centering
    \includegraphics[width=\linewidth]{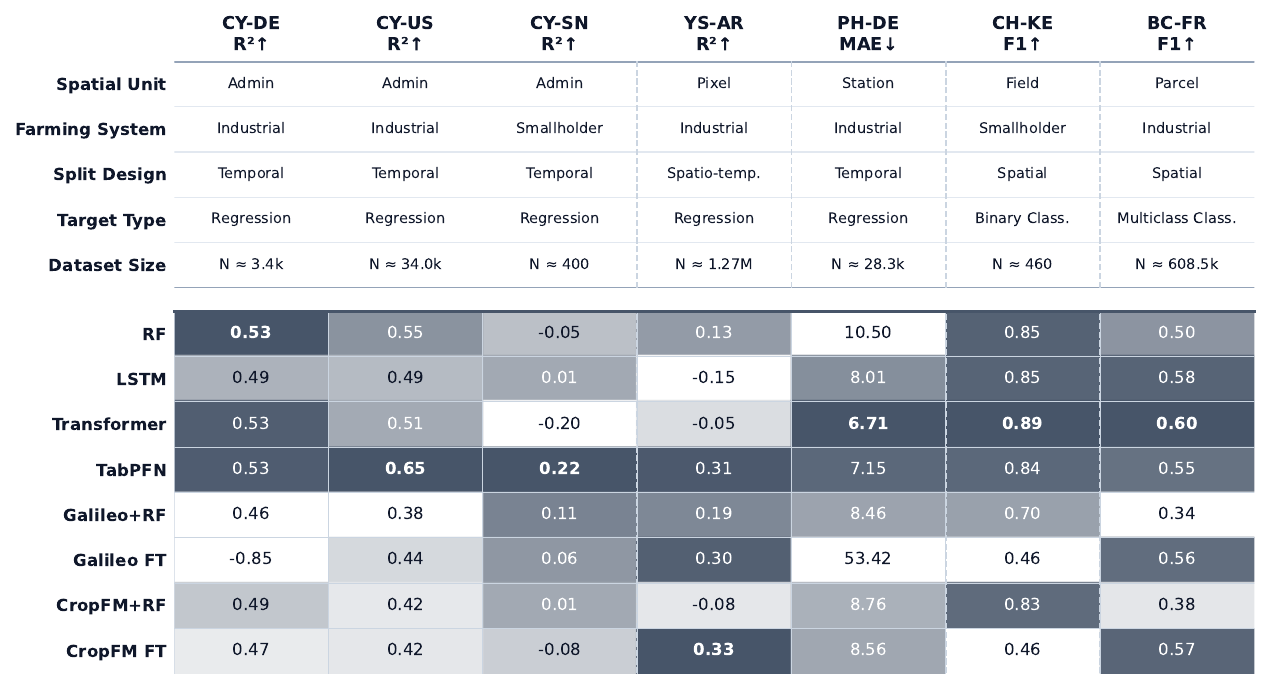}
    \caption{\textbf{Task Heterogeneity and Performance Benchmarking.} The top panel maps the structural axes (spatial unit, farming system, split design, task type, and dataset size) characterizing each dataset. The bottom heatmap displays model performance across these heterogeneous targets; darker shades indicate better relative performance after per-dataset normalization, and bold values denote the best-performing model per column. No single architecture consistently dominates, highlighting the volatility induced by spatial-temporal shifts. Abbreviations: CY-DE/US/SN (CY-Bench - Germany, USA, Senegal), YS-AR (YieldSAT - Argentina), PH-DE (Phenology - Germany), CH-KE (CropHarvest - Kenya), BC-FR (BreizhCrop - France); FT denotes full fine-tuning.}
    \label{fig:performance_heatmap}
\end{figure*}

\subsection{Challenge 2: Task Heterogeneity}
\label{sec:task_heterogeneity}

\subsubsection{\textbf{Structural Heterogeneity of the Task Suite}}
To formalize the scale of this challenge, we categorize our downstream datasets across five distinct "structural axes" of variation, detailed in the top panel of Figure \ref{fig:performance_heatmap}. These axes define the massive environmental and geometric shifts that a FM must bridge:

\begin{itemize}[nosep,leftmargin=0.35cm, labelsep=0.15cm]
    \item \textbf{Spatial and Resolution Heterogeneity:} Although all datasets provide point-wise features for specific geographic locations, what these individual points represent varies drastically. This ranges from raw 10m pixels (in YieldSAT, CropHarvest, and BreizhCrop) and point-station level observation sites (in phenology), to county-level aggregated regions (in CY-Bench). This structural mismatch forces foundation models (which are typically pre-trained on uniform, high-resolution pixel grids e.g., 10m) to adapt to highly varied spatial aggregation levels. Furthermore, even within datasets sharing similar native pixel resolutions (e.g., CropHarvest vs. BreizhCrop), the underlying landscapes represent contrasting agricultural paradigms, ranging from fragmented, smallholder plots in Kenya to large-scale, industrial fields in France \cite{fao_book}.
    
    \item \textbf{Temporal Dynamics and Evaluation:} The temporal evaluation targets shift from point-in-time classification (e.g., predicting crop types at the end of a season) to continuous, highly non-linear regression profiles. In phenology estimation, the target variable is a physical day-of-year event driven by cumulative thermal curves. Conversely, yield estimation requires mapping a complex, year-long sequence of weather and soil signals to a single scalar mass value. Furthermore, the temporal framing can also differ, namely middle-of-season and end-of-season estimation of yield.
    
    \item \textbf{Label Density and Quality:} The datasets transition from dense, continental-scale, and clean administrative registries (e.g., official agricultural records in Germany and the US) to highly sparse, noisy, and manually collected datasets in underrepresented regions like Senegal and Kenya. In these latter data-scarce regimes, severe class imbalances introduce additional noise that compounds the spatial-temporal shift.
\end{itemize}

\subsubsection{\textbf{Rank Instability Across Datasets}}
The direct consequence of this structural heterogeneity is a lack of consistency in model performance. Rather than showing a steady hierarchy of model capability, the performance heatmap (Figure \ref{fig:performance_heatmap}) reveals a highly volatile landscape where the relative order of models fluctuates between datasets. On highly structured, data-rich yield estimation tasks like \texttt{CY-US}, the unconstrained, tabular-focused TabPFN outperforms all other models with an $R^2$ of $0.65$. However, on the Argentinian yield dataset (\texttt{YS-AR}), the hierarchy completely flips: CropFM under full fine-tuning yields the best performance ($R^2 = 0.33$), while sequential baselines like LSTMs suffer a complete collapse ($R^2 = -0.15$). On crop type mapping tasks (\texttt{CH-KE} and \texttt{BC-FR}), the sequence-based Transformer performs better than the frozen FM features, yielding the top F1-scores ($0.89$ and $0.60$, respectively). 

\subsubsection{\textbf{The Risk of Finetuning}} We note that the results of finetuning Galileo on German yield estimation (CY-DE) and the German phenology estimation (PH-DE) yield very poor performance ($R^2 = -0.85$, and 53.42 MAE). We hypothesize that this is due to catastrophic forgetting of pretrained representations when finetuning an overparameterized model on a localized small-scale dataset under shifts \cite{kumar2022finetuningdistortpretrainedfeatures}.

\section{Reflections on Agricultural FM Development}
\label{sec:reflections}
Agricultural remote sensing presents structural challenges that distinguish it from the domains in which FMs have achieved their most celebrated successes \cite{tan2023promiseschallengesmultimodalfoundation}. Our evaluation surfaces three properties of agricultural monitoring that are fundamental to the problem domain, not incidental implementation difficulties, and one practical barrier the community has not yet adequately addressed. Together, they suggest that adapting FMs to agriculture requires more than scaling existing approaches.

\paragraph{\textbf{The Modality Superset Problem}}
FM transfer assumes pretraining and downstream tasks share a compatible input space. This assumption breaks down in agricultural monitoring in a way with no direct analogue in NLP or computer vision: different agricultural datasets provide different subsets of modalities, whichever the dataset authors considered most relevant to their task, while FMs are pretrained on diverse but fixed multi-modal data \cite{tseng2024lightweightpretrainedtransformersremote, tseng2025galileolearningglobal}. This makes it difficult to build a general-purpose agricultural FM spanning a superset of every modality any dataset might need, and no single downstream task provides the full modality set simultaneously: crop type mapping provides only optical imagery, sub-national yield estimation relies on aggregated weather and vegetation indicators, pixel-level yield estimation offers a richer set, and phenology reduces to a single meteorological variable. A deployed model therefore always operates on a strict subset of its training distribution, with no mechanism to signal which modalities are absent rather than merely uninformative. Our results show this is not uniformly harmful: learned representations partially compensate for missing modalities in some settings, but the compensation is unreliable and dataset-dependent in ways that are difficult to predict.

\paragraph{\textbf{Task Heterogeneity as a Fundamental Barrier}}
The breadth of "agricultural monitoring" poses a challenge model design alone cannot resolve. Our task suite spans pixel-level and sub-national yield estimation, point-observation phenology, and field- and parcel-level crop classification, tasks that differ not only in output target but in spatial granularity, temporal aggregation, label provenance, label density, and farming system. The rank instability we observe, where no model ranks consistently across datasets, reflects this structural diversity rather than any individual model's failure, consistent with findings in general EO benchmarks \cite{marsocci2025pangaeaglobalinclusivebenchmark, simumba2026geobench2performancecapabilityrethinking}. No single pretraining objective optimizes for all of these settings: a reconstruction objective over pixel time series suits phenological and fine-grained spectral tasks, but loses relevance once targets are aggregated to county-level annual scalars. Meaningful progress requires embracing this heterogeneity as a first-class property of the domain, designing for it explicitly rather than benchmarking around it.

\paragraph{\textbf{Ease of Use and the Practical Deployment Gap}}
Performance benchmarks rarely capture the practical cost of deployment. Deploying an FM requires choosing between frozen feature extraction (a Random Forest or linear head over the encoder) and full fine-tuning where labels permit; the former requires selecting and tuning a probe architecture, the latter demands careful regularization to avoid the representation collapse we observe in several experiments, and both are compounded when a task cannot supply the modalities the encoder expects. Tabular baselines such as TabPFN and Random Forest require none of this: they ingest whatever modalities are available with minimal preprocessing, and in our evaluation are competitive with or superior to FMs in multiple settings. This does not argue against agricultural FM development, but it does show that the practical gap between deploying an FM and a well-tuned tabular baseline currently exceeds the performance gap in many agricultural monitoring contexts.

\section{Conclusion}

We evaluate a diverse set of foundation models and standard baselines across three agricultural monitoring tasks central to food security, asking not merely whether these models generalize but, when they fail, why. Rather than treating agriculture as one more downstream domain for a general-purpose EO backbone, we construct a evaluation suite with seven datasets spanning the axes along which agricultural monitoring actually varies, spatial unit, farming system, split design, target type and dataset size to isolate the conditions under which FMs break down. Across pixel-level crop type mapping, point-observation phenology, and administratively aggregated yield estimation over industrial and smallholder systems, failures trace back to specific structural mismatches between how these models are built and what agricultural monitoring demands: FMs pay a substantial modality-mismatch penalty whenever a dataset depends on variables outside their pretraining set, and even under full modality coverage, frozen representations do not consistently outperform simple, unconstrained baselines. No single model ranks consistently across datasets; the best architecture changes with geography, spatial aggregation, and label regime, underscoring that no single design choice, pretraining objective, adaptation strategy, or architecture, is sufficient on its own.

These findings motivate the three challenges above: the modality superset problem, the fundamental heterogeneity of agricultural tasks, and the deployment gap separating FMs from well-tuned tabular baselines. None of this argues against agricultural foundation models as a direction; our results with CropFM suggest models designed explicitly around agriculture's modality and resolution requirements can close part of the gap. It argues instead for a shift in how the community builds and evaluates them: architectures that ingest varied modalities, benchmarks that treat task heterogeneity as a first-class evaluation axis rather than averaging it away, and honesty about when a simple supervised baseline is the right tool for the job. As these models increasingly inform decisions with real consequences for farmers, closing this gap responsibly will also require looking beyond the industrial systems most current data comes from, toward the smallholder, data-scarce systems where the stakes of getting it wrong are highest.

\section{Limitations and Ethical Considerations}

Several limitations exist within this research that should be noted. First, our evaluation relies on EO models, but not many EO models can be structurally applied to these datasets, as they are not natively designed for long temporal sequences or built to ingest a variety of modalities. This slightly limits the scope of cross-comparisons between various models. Secondly, while the evaluation tests a variety of structural shifts, it does not consider a truly global scale of different regions, diverse farming practices, and other relevant agricultural tasks. Thirdly, the study utilizes traditional, simple methods as baselines, whereas a large body of literature exists detailing complex, domain-specific models designed for each individual task. Finally, the most critical, while this study highlights two critical challenges, there could also be other unknown challenges that cause such rank instabilities across datasets, for example, there could be some biases in what FMs were pretrained on and information leakage to the downstream tasks.

From an ethical standpoint, the deployment of FMs in agriculture introduces several significant risks. It is critical that pre-training corpora avoid systemic bias toward data-rich, large-scale industrial farming systems, which are predominantly located in the US and Europe. Because these models can inform high-stakes applications such as crop insurance estimation and regional food security planning, biased representations can actively distort local market prices and systematically disadvantage vulnerable farming communities. Therefore, these representational biases must be strictly considered when developing general-purpose models for agriculture.

\section{Generative AI Usage} 

Generative AI tools were used in this work primarily to assist with code development and manuscript drafting, specifically Google Antigravity, Gemini and Claude. Following the use of these tools, the code and the text were double checked, validated and also modified when necessary to ensure technical accuracy and scientific integrity. The authors take full responsibility for the final content of this research. 

\section{Data and Code Availability}

To support full reproducibility, the benchmarking codebase for all downstream tasks is open-sourced at \url{https://github.com/vishalned/CropBench}. All data utilized in this benchmark is sourced from existing, publicly available datasets. Although the original respective GitHub repositories provide their own dataloaders, we have unified them into a single, standardized pipeline to streamline the evaluation of diverse models. 

\begin{acks}
This work was supported by the EU Horizon project "AI Foundation Models in Agricultural Sciences" (grant number 101293777). We further acknowledge the Dutch national e-infrastructure, provided with the support of SURF Cooperative under grant no. EINF-15689. We also thank other members part of the Artificial Intelligence Group at Wageningen University and Research who gave feedback with the writing of this manuscript.
\end{acks}

\newpage
\bibliographystyle{ACM-Reference-Format}
\bibliography{sample-base}

\appendix
\section{Granular Modality and Feature Utilization}

To accurately compute the modality coverage percentages, we calculate the ratio at the granular feature level rather than by high-level modality group. Table \ref{tab:granular_modalities} details the exact variables utilized by each downstream task and the native ingestion capabilities of the evaluated foundation models.

\begin{table*}
    \centering
    \caption{Granular feature breakdown across tasks and foundation model native support. Acronyms: Digital Elevation Model (DEM), Average/Maximum/Minimum Temperature ($t_{avg}$, $t_{max}$, $t_{min}$), Precipitation (prec), Climate Water Balance (cwb), Solar Radiation (rad), Fraction of Absorbed Photosynthetically Active Radiation (fapar), Normalized Difference Vegetation Index (ndvi), Surface Soil Moisture (ssm), Available Water Capacity (awc), Elevation (elev), Cation Exchange Capacity (cec), Coarse Fragments Volumetric (cfvo), Soil Organic Carbon (soc).}
    \label{tab:granular_modalities}
    \resizebox{\textwidth}{!}{
    \begin{tabular}{@{}llp{6cm}ccc@{}}
        \toprule
        \textbf{Dataset} & \textbf{Category} & \textbf{Specific Variables (Features)} & \textbf{Count} & \textbf{Galileo} & \textbf{CropFM} \\
        \midrule
        \multirow{2}{*}{\textbf{BreizhCrop}} 
        & Satellite & Sentinel-2 (10 bands) & 10 & 10 & 10 \\
        & \textit{Total} & & \textit{10} & \textit{10 (100\%)} & \textit{10 (100\%)} \\
        \midrule
        \multirow{2}{*}{\textbf{Phenology}} 
        & Meteo & $t_{avg}$ & 1 & 1 & 1 \\
        & \textit{Total} & & \textit{1} & \textit{1 (100\%)} & \textit{1 (100\%)} \\
        \midrule
        \multirow{5}{*}{\textbf{CropHarvest}} 
        & Satellite & Sentinel-1 (2), Sentinel-2 (10) & 12 & 12 & 12 \\
        & Meteo & $t_{avg}$, prec & 2 & 2 & 2 \\
        & DEM & elev, slope & 2 & 2 & 2 \\
        & Bio/Hydro & ndvi & 1 & 1 & 0 \\
        & \textit{Total} & & \textit{17} & \textit{17 (100\%)} & \textit{16 (94.1\%)} \\
        \midrule
        \multirow{5}{*}{\textbf{CY-Bench}} 
        & Meteo & $t_{min}, t_{max}, t_{avg}$, prec, cwb, rad & 6 & 2 & 5 \\
        & Bio/Hydro & fapar, ndvi, ssm & 3 & 2 & 1 \\
        & Soil & awc, bulk\_density & 2 & 0 & 0 \\
        & \textit{Total} & & \textit{11} & \textit{4 (36.4\%)} & \textit{6 (54.5\%)} \\
        \midrule
        \multirow{5}{*}{\textbf{YieldSAT}} 
        & Satellite & Sentinel-2 (10 bands) & 10 & 10 & 10 \\
        & Meteo & $t_{avg}, t_{max}, t_{min}$, prec & 4 & 2 & 4 \\
        & DEM & elev, slope & 2 & 2 & 2 \\
        & Soil & cec, cfvo, clay, nitrogen, phh2o, sand, silt, soc (each at 6 depths) & 48 & 0 & 12 \\
        & \textit{Total} & & \textit{64} & \textit{14 (21.9\%)} & \textit{30 (46.9\%)} \\
        \bottomrule
    \end{tabular}
    }
\end{table*}

\section{Full set of results} 
\label{sec:app_full_results}

We provide the complete set of results in Table \ref{tab:app_full_results}. An additional experiment was further conducted to run the supervised baselines using similar constrainsts as CropFM i.e using the same set of modalities available to CropFM. These results are reported in \ref{tab:app_cropfm_comparison}.

\begin{table*}
\centering
\caption{Main results across all datasets and models, evaluated at each model's native modality access (i.e., all available features for supervised baselines and TabPFN; natively supported modalities for Galileo and CropFM). For EO FMs, \textit{Frozen} denotes a Random Forest trained on frozen embeddings and \textit{FT} denotes full fine-tuning. Metrics are $R^2$ for yield estimation, F1-score for crop classification, and MAE (days) for phenology; bold values denote the best-performing model per column. Results are averaged over three seeds, $\pm$ one standard deviation.}
\label{tab:app_full_results}
\resizebox{\textwidth}{!}{
\begin{tabular}{l cccc cc c}
\toprule
 & \multicolumn{4}{c}{\textbf{Yield Estimation ($R^2$) $\uparrow$}} & \multicolumn{2}{c}{\textbf{Crop Classification (F1) $\uparrow$}} & \textbf{Phenology (MAE) $\downarrow$} \\
\cmidrule(lr){2-5} \cmidrule(lr){6-7} \cmidrule(lr){8-8}
\textbf{Model} & \textbf{CY-US} & \textbf{CY-SN} & \textbf{CY-DE} & \textbf{YS-AR} & \textbf{BC-FR} & \textbf{CH-KE} & \textbf{PH-DE} \\
\midrule
RF                  & 0.55 $\pm$ 0.00 & -0.05 $\pm$ 0.00 & \textbf{0.53 $\pm$ 0.00} & 0.13 $\pm$ 0.01 & 0.50 $\pm$ 0.00 & 0.85 $\pm$ 0.01 & 10.50 $\pm$ 0.05 \\
LSTM                & 0.49 $\pm$ 0.03 & 0.01 $\pm$ 0.04  & 0.49 $\pm$ 0.06 & -0.15 $\pm$ 0.06 & 0.58 $\pm$ 0.00 & 0.85 $\pm$ 0.03 & 8.01 $\pm$ 0.03 \\
Transformer         & 0.51 $\pm$ 0.03 & -0.20 $\pm$ 0.09 & 0.53 $\pm$ 0.02 & -0.05 $\pm$ 0.21 & \textbf{0.60 $\pm$ 0.00} & \textbf{0.89 $\pm$ 0.04} & \textbf{6.71 $\pm$ 0.12} \\
TabPFN              & \textbf{0.65 $\pm$ 0.00} & \textbf{0.22 $\pm$ 0.02} & 0.53 $\pm$ 0.00 & 0.31 $\pm$ 0.01 & 0.55 $\pm$ 0.00 & 0.84 $\pm$ 0.00 & 7.15 $\pm$ 0.06 \\
\addlinespace
Galileo (Frozen)    & 0.38 $\pm$ 0.00 & 0.11 $\pm$ 0.01 & 0.46 $\pm$ 0.00 & 0.19 $\pm$ 0.00 & 0.34 $\pm$ 0.00 & 0.70 $\pm$ 0.01 & 8.46 $\pm$ 0.05 \\
Galileo (FT)        & 0.44 $\pm$ 0.01 & 0.06 $\pm$ 0.07 & -0.85 $\pm$ 0.15 & 0.30 $\pm$ 0.04 & 0.56 $\pm$ 0.00 & 0.46 $\pm$ 0.00 & 53.42 $\pm$ 0.21 \\
\addlinespace
CropFM (Frozen)     & 0.42 $\pm$ 0.00 & 0.01 $\pm$ 0.04 & 0.49 $\pm$ 0.00 & -0.08 $\pm$ 0.01 & 0.38 $\pm$ 0.00 & 0.83 $\pm$ 0.02 & 8.76 $\pm$ 0.08 \\
CropFM (FT)         & 0.42 $\pm$ 0.01 & -0.08 $\pm$ 0.04 & 0.47 $\pm$ 0.03 & \textbf{0.33 $\pm$ 0.04} & 0.57 $\pm$ 0.01 & 0.46 $\pm$ 0.00 & 8.56 $\pm$ 0.08 \\
\bottomrule
\end{tabular}
}
\end{table*}

\begin{table*}
\centering
\caption{CropFM-modality comparison. To isolate the contribution of CropFM's wider native modality access from its pretrained priors, we re-evaluate the supervised baselines restricted to the same modality subset available to CropFM, and compare them directly against CropFM's frozen and fully fine-tuned variants.}
\label{tab:app_cropfm_comparison}
\resizebox{\textwidth}{!}{
\begin{tabular}{l cccc cc c}
\toprule
 & \multicolumn{4}{c}{\textbf{Yield Estimation ($R^2$) $\uparrow$}} & \multicolumn{2}{c}{\textbf{Crop Type Mapping (F1) $\uparrow$}} & \textbf{Phenology (MAE) $\downarrow$} \\
\cmidrule(lr){2-5} \cmidrule(lr){6-7} \cmidrule(lr){8-8}
\textbf{Model} & \textbf{CY-US} & \textbf{CY-SN} & \textbf{CY-DE} & \textbf{YS-AR} & \textbf{BC-FR} & \textbf{CH-KE} & \textbf{PH-DE} \\
\midrule
RF (CropFM modalities)          & 0.49 $\pm$ 0.00 & 0.00 $\pm$ 0.02 & 0.13 $\pm$ 0.00 & 0.07 $\pm$ 0.01 & 0.50 $\pm$ 0.00 & 0.81 $\pm$ 0.02 & 10.50 $\pm$ 0.05 \\
LSTM (CropFM modalities)        & 0.44 $\pm$ 0.03 & -0.14 $\pm$ 0.14 & 0.35 $\pm$ 0.08 & 0.28 $\pm$ 0.04 & 0.58 $\pm$ 0.00 & 0.68 $\pm$ 0.03 & 8.01 $\pm$ 0.03 \\
Transformer (CropFM modalities) & \textbf{0.52 $\pm$ 0.02} & -0.07 $\pm$ 0.12 & 0.47 $\pm$ 0.05 & 0.08 $\pm$ 0.11 & \textbf{0.60 $\pm$ 0.00} & \textbf{0.87 $\pm$ 0.02} & \textbf{6.71 $\pm$ 0.12} \\
\addlinespace
CropFM (Frozen)                 & 0.42 $\pm$ 0.00 & \textbf{0.01 $\pm$ 0.04} & \textbf{0.49 $\pm$ 0.00} & -0.08 $\pm$ 0.01 & 0.38 $\pm$ 0.00 & 0.83 $\pm$ 0.02 & 8.76 $\pm$ 0.08 \\
CropFM (FT)                     & 0.42 $\pm$ 0.01 & -0.08 $\pm$ 0.04 & 0.47 $\pm$ 0.03 & \textbf{0.33 $\pm$ 0.04} & 0.57 $\pm$ 0.01 & 0.46 $\pm$ 0.00 & 8.56 $\pm$ 0.08 \\
\bottomrule
\end{tabular}
}
\end{table*}

\section{Building CropFM}

\subsection{Pretraining Data}
\subsubsection{Spatial Sampling}
We restrict pretraining to regions with active temporary-crop agriculture as identified by the ESA WorldCereal Google Earth Engine (GEE) product. For each country in the world, we tile the country's land area into a regular 5km~$\times$5km grid. A grid cell is then retained if it contains atleast 5km$^2$ of WorldCereal temporary cropland; all other cells are discarded. Within each retained cell, we choose 10 points per grid via stratified random sampling, restricted to pixels classified as temporary cropland (rather than uniformly sampling the whole cell). In total we sample $\approx 1 Million$ locations.

\subsubsection{Modalities}
For every sampled point, we query GEE for a fixed one-year window (2021-01-01 to 2022-01-01). Nine modalities are used, split into temporal and static groups, summarized in Table~\ref{tab:app_modalities}. For each temporal modality, we download all the data within the one-year window.

\begin{table*}
\caption{Optimization and training configuration for CropFM pretraining.}
\label{tab:training-config}
\resizebox{0.5\textwidth}{!}{
\begin{tabular}{ll}
\toprule
Parameter & Value \\
\midrule
Optimizer & AdamW ($\beta_1=0.9$, $\beta_2=0.95$) \\
Weight decay & 0.05 \\
Learning rate & $1.5 \times 10^{-4}$ \\
LR schedule & Cosine annealing, $\eta_{\min} = 1 \times 10^{-6}$ \\
Batch size & 256 \\
Gradient accumulation & $\times 8$ (effective batch size 2048) \\
Epochs & up to 40 \\
Hardware & 1$\times$ NVIDIA A100 (80 GB) \\
Train/val/test split & 80\% / 10\% / 10\%, contiguous index ranges \\
\bottomrule
\end{tabular}
}
\end{table*}

\subsubsection{Temporal Harmonization}

Since different sensors revisit at very different periods, the temporal modalities are resampled onto a common weekly grid (mean aggregation by default), truncated to 52 weeks. We also encode the coordinates and the calender position of the week, month. 

\subsubsection{Normalization statistics}
Normalization statistics are computed per variable of every modality separately and using only non-Nan entries. 

\subsection{Model}
Following Presto \cite{tseng2024lightweightpretrainedtransformersremote}, CropFM uses a single generic lightweight transformer, where every modality, timestep pair is projected to one token via a modality-specific linear layer and all tokens share one transformer encoder/decoder. 

\begin{itemize}
    \item \textbf{Tokenization}: each modality has its own $\text{Linear}(\text{in\_dim} \rightarrow d_{\text{model}})$ tokenizer (in\_dim = number of variables for that modality, e.g.\ 8 for AgERA5, 12 for Sentinel-2). Static modalities contribute a single token. Timesteps flagged invalid are replaced with a learnable per-modality ``missing'' token, so the model can distinguish a genuine data gap (e.g., no cloud-free Sentinel-2 scene that week) from a token deliberately withheld by MAE masking.
    \item \textbf{Positional information}: every token receives (i) a fixed sinusoidal positional encoding by sequence position (zeroed out for static tokens), (ii) a fixed sinusoidal week-of-year encoding for temporal tokens only, and (iii) a learnable embedding identifying its modality.
    \item \textbf{Encoder and Decoder}: depth 2 standard transformer blocks (similar to Presto)
\end{itemize}

\subsubsection{Masking Strategy}
\label{sec:masking-strategy}
 
We use a structured masking scheme rather than pure random masking. On every forward pass, one of four strategies is drawn uniformly at random (equal 25\% weight each):
 
\begin{enumerate}
    \item \textbf{Random}: tokens masked independently across the whole sequence.
    \item \textbf{Modality}: entire modalities are masked ($\sim$45\% of the 9 modalities per sample), forcing reconstruction of one modality purely from the others.
    \item \textbf{Random timesteps}: within each modality independently, $\sim$75\% of timesteps are masked, keeping cross-modality structure at each timestep intact.
    \item \textbf{Contiguous / seasonal}: only timesteps matching a fixed calendar pattern are kept (e.g., one sample per month, or only the first/second half of the year), forcing longer-range temporal interpolation/extrapolation.
\end{enumerate}

\subsubsection{Optimization and Training Configuration}

Table~\ref{tab:training-config} summarizes the optimization and training setup.

\begin{table*}
\centering
\footnotesize
\caption{Data modalities used for CropFM pretraining. All sources are Google Earth Engine (GEE) asset IDs unless noted otherwise.}
\label{tab:app_modalities}
\begin{tabularx}{\textwidth}{@{} l L l L @{}}
\toprule
\multicolumn{4}{l}{\textbf{Temporal}} \\
\midrule
Modality & Source (GEE asset ID) & Native res. & Variables used \\
\midrule
Sentinel-2 L2A & \texttt{COPERNICUS/S2\_SR\_HARMONIZED} & 10 m & 12 spectral bands (B1--B9, B11, B12); scenes filtered to $<$50\% cloud cover \\
Sentinel-1 GRD & \texttt{COPERNICUS/S1\_GRD} & 10 m & VV, VH \\
AgERA5 & \texttt{projects/climate-engine-pro/assets/ce-ag-era5-v2/daily} & $\sim$10 km & precipitation, vapour pressure, min/mean/max 2 m air temperature, solar radiation, wind speed, max relative humidity (8 vars) \\
FAPAR/LAI & \texttt{NOAA/CDR/VIIRS/LAI\_FAPAR/V1} & $\sim$5.6 km & FAPAR, LAI \\
\midrule
\multicolumn{4}{l}{\textbf{Static}} \\
\midrule
Modality & Source (GEE asset ID) & Native res. & Variables used \\
\midrule
Soil & \texttt{projects/soilgrids-isric/\{clay,nitrogen,phh2o,soc\}\_mean} & 250 m & clay, nitrogen, pH, SOC; each at 3 depths (0--5, 5--15, 15--30 cm) \\
Elevation & \texttt{projects/sat-io/open-datasets/ASTER/GDEM} & 10 m & elevation, slope \\
Coordinates & --- (computed, not a GEE asset) & --- & cyclic sin/cos encoding of lon/lat \\
WorldCereal crop mask & \texttt{ESA/WorldCereal/AEZ/v100} + AEZ table & --- & AEZ id, crop class (none / maize / winter cereals / spring cereals) \\
WorldCereal crop calendar & \texttt{ESA/WorldCereal/AEZ/v100} & --- & start/end of season for maize, winter cereals, spring cereals \\
\bottomrule
\end{tabularx}
\end{table*}

\end{document}